\documentclass[letterpaper]{article} % DO NOT CHANGE THIS
\usepackage{aaai2026}  % DO NOT CHANGE THIS
\usepackage{times}  % DO NOT CHANGE THIS
\usepackage{helvet}  % DO NOT CHANGE THIS
\usepackage{courier}  % DO NOT CHANGE THIS
\usepackage[hyphens]{url}  % DO NOT CHANGE THIS
\usepackage{graphicx} % DO NOT CHANGE THIS
\usepackage{subcaption}
\usepackage{booktabs}
\usepackage{natbib}  % DO NOT CHANGE THIS AND DO NOT ADD ANY OPTIONS TO IT
\usepackage{caption} % DO NOT CHANGE THIS AND DO NOT ADD ANY OPTIONS TO IT
\usepackage[utf8]{inputenc}
\usepackage[T1]{fontenc}
\usepackage{xcolor}
\usepackage{enumitem}
\usepackage{adjustbox}
\usepackage{float}
\usepackage{amsmath}
\usepackage{amssymb}
\usepackage{amsfonts}
\usepackage{mathtools}
\usepackage{multirow, makecell}
\usepackage{bm}
\usepackage{amsthm}
\usepackage{tabularx}
\usepackage{booktabs}
\usepackage{array}
\usepackage{tcolorbox}
\usepackage{url}
\usepackage{algorithm}
\usepackage{algpseudocode}
\usepackage{framed}
\usepackage{dsfont}

\newcommand{\href}[2]{#2} % 显示文本，忽略链接
\newcommand{\BibUrl}[1]{} % 忽略BibUrl
\newcommand{\BibAnnote}[1]{} % 忽略BibAnnote

\title{DeepJSONEval: Benchmarking Complex Nested JSON Data Mining for Large Language Models}

\author{
	Zhicheng Zhou\equalcontrib,
	Jing Li\equalcontrib,
	Suming Qiu, Junjie Huang, Linyuan Qiu, Zhijie Sun\thanks{Corresponding author.}
}

\nocopyright 

\affiliations{
	\textsuperscript{\rm 1}GTS, Huawei Technologies Co., Ltd\\
	Correspondence to: sunzhijie3@huawei.com
}

\begin{document}
	
	\maketitle

	\begin{abstract}
		The internet is saturated with low-density, high-redundancy information, such as social media comments, repetitive news, and lengthy discussions, making it difficult to extract valuable insights efficiently. Multi-layer nested JSON structures provide an effective solution by compressing such information into semantically rich, hierarchical representations, which organize data into key-value pairs, arrays, and nested objects, preserving contextual relationships and enabling efficient storage, retrieval, and semantic querying. For instance, in news aggregation, a JSON object can nest an article's metadata (title, author, date), content (text, multimedia), and multimedia information (multimedia type, caption) hierarchically. Large Language Models (LLMs) play a transformative role in web data mining by parsing unstructured text and outputting structured results directly into complex JSON schemas. However, current benchmarks for evaluating LLMs' JSON output capabilities overemphasize pure JSON generation rather than assessing data comprehension and extraction abilities, a limitation that lacks relevance to practical web data mining tasks. To address this, we introduce DeepJSONEval, a novel benchmark featuring 2100 multi-domain instances with deep nested structures, categorized by difficulty. Experiments show significant performance gaps among LLMs in handling such complexity. Our benchmark and datasets are open-sourced to advance research in structured JSON generation. (\href{}{https://github.com/GTS-AI-Infra-Lab-SotaS/DeepJSONEval}).
	\end{abstract}
	
%	\begin{CCSXML}
%		<ccs2012>
%		<concept>
%		<concept_id>10002951.10003260.10003277</concept_id>
%		<concept_desc>Information systems~Web mining</concept_desc>
%		<concept_significance>300</concept_significance>
%		</concept>
%		</ccs2012>
%	\end{CCSXML}
%	
%	\ccsdesc[300]{Information systems~Web mining}
%	
%	
%	%%
%	%% Keywords. The author(s) should pick words that accurately describe
%	%% the work being presented. Separate the keywords with commas.
%	\keywords{Multi-layer Nested JSON, Web Mining, Large Language Models (LLMs), Benchmark}
%	%% A "teaser" image appears between the author and affiliation
%	%% information and the body of the document, and typically spans the
%	%% page.
%	
%	
%	%%
%	%% This command processes the author and affiliation and title
%	%% information and builds the first part of the formatted document.

	\section{Introduction}
	The exponential growth of digital content has created an information extraction paradox: while data volume increases dramatically, information density remains critically low due to redundant social media posts, repetitive news coverage, and poorly structured web content. This sparsity-abundance contradiction demands robust information condensation techniques that can transform noisy, heterogeneous data into structured, semantically rich representations suitable for computational analysis.
	
	Multi-layer nested JSON structures have emerged as a powerful solution, enabling hierarchical compression of sparse information through systematic key-value organization and nested object relationships. Unlike flat data formats, nested JSON preserves complex semantic dependencies while maintaining both machine readability and human interpretability—essential for capturing nuanced information relationships in domains ranging from news aggregation (article metadata nested with content analysis and multimedia details) to financial analytics (stock data with nested performance metrics and temporal indicators)\cite{syafiq2025cityjson,chinta2025json}. 
	
\begin{figure*}[t]
		\centering
		\includegraphics[width=0.9\textwidth]{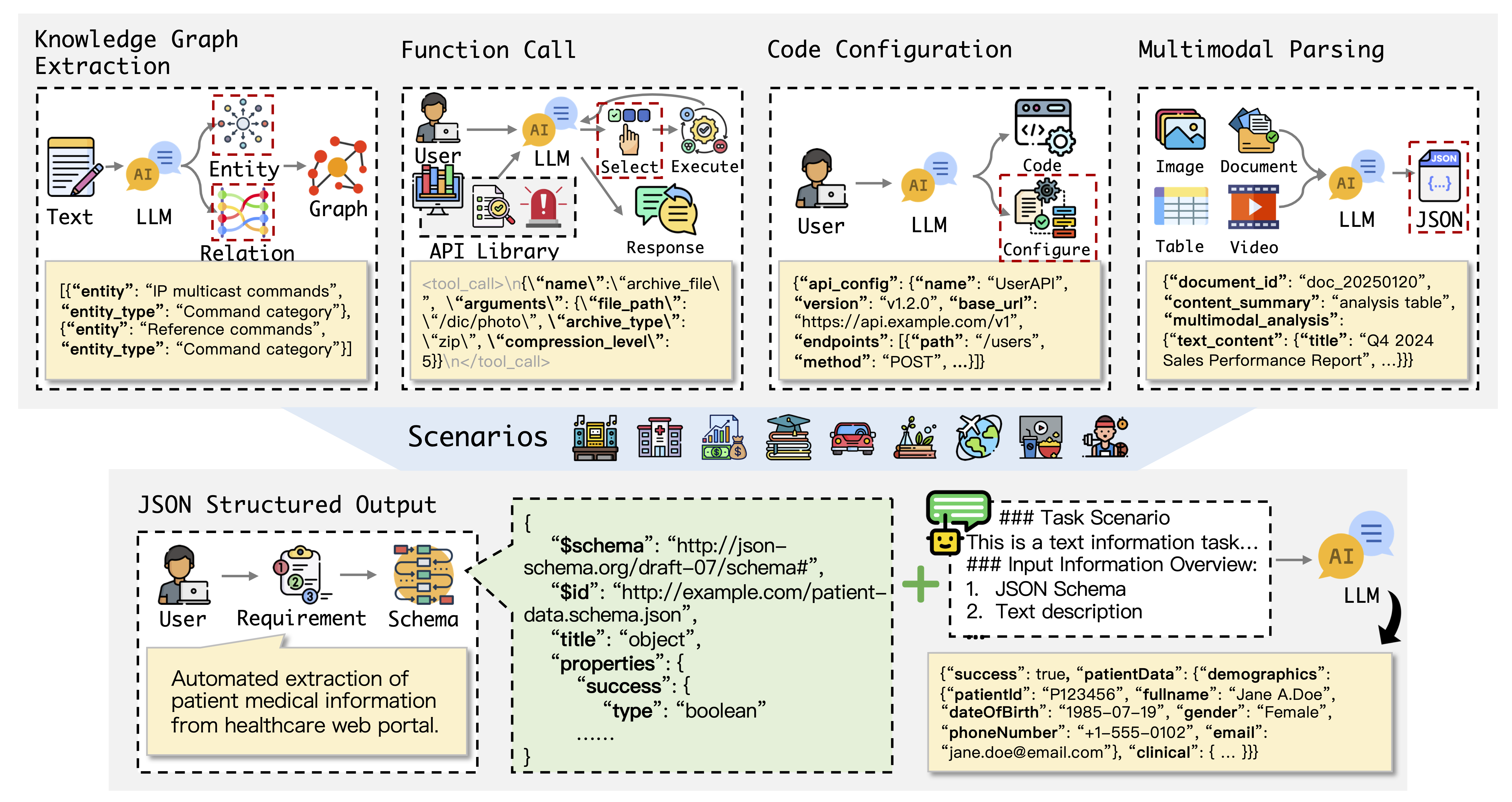}
		\caption{The representative application scenarios for multi-layer nested JSON with LLM.}
		\label{scenarios}
	\end{figure*}

	Recent advancements in artificial intelligence, particularly the emergence of Large Language Models (LLMs), like GPT-4, have become transformative agents in web mining and content analysis, fundamentally altering how people extract, interpret, and structure information from the vast and often chaotic digital corpus \cite{guo2025structuredoutputsenablegeneralpurpose}.LLMs' capabilities in natural language understanding, content summarization, entity identification, and relationship inference allow them to convert raw, noisy web data, such as news articles, forum discussions, and product reviews into structured, actionable JSON (see Figure~\ref{scenarios}) \cite{xu2025chatpd}.

	However, evaluating these capabilities presents significant challenges, as traditional benchmarks often fail to capture the complexity and nuanced requirements of extracting key information from information-saprse web data and convert the key information into multi-lavel nested JSON structrue. Current benchmarks lack standardized evaluation of multi-layer JSON generation quality. This gap necessitates a comprehensive evaluation dataset that can systematically measure the fidelity, completeness, and structural appropriateness of JSON-based information extraction systems, moving beyond traditional metrics that fail to capture the full spectrum of challenges in web data mining applications.
	
	While existing evaluation frameworks have made notable advances in assessing LLM capabilities across various dimensions, including contamination-resistant evaluation\cite{white2024livebenchachallengingcontaminationlimitedllm}, instruction-following verification\cite{zhou2023instructionfollowingevaluationlargelanguage}, mobile function calling\cite{wang2024hammerbenchfinegrainedfunctioncallingevaluationin}, schema-based information extraction\cite{gui2024iepileunearthinglargescaleschemabased, geng2025jsonschemabencharigorousbenchmarkof, yang2025structeval}. Systematic analysis reveals fundamental gaps that limit their applicability to real-world JSON-based information extraction.

	A fundamental conceptual distinction emerges: existing benchmarks \cite{yang2025structeval, LIU2024103809, xia2024fofobenchmarkevaluatellms, geng2025jsonschemabencharigorousbenchmarkof} predominantly frame JSON generation as schema-adherence tasks, wherein models synthesize structures from templates without source content, providing limited evaluation of information extraction from unstructured text—essential for web data mining applications.

	Current frameworks exhibit constraining characteristics: monolingual focus (predominantly English), limited domain diversity (exceptions: SoEval \cite{LIU2024103809}, Schema-Guided Dialogue \cite{rastogi2020scalablemultidomainconversationalagents}), variable structural complexity (shallow nesting in json-mode-eval \cite{nousresearch_json-mode-eval}, STRUCTUREDRAG \cite{shorten2024structuredragjsonresponseformatting}, NESTFUL \cite{basu2025nestfulbenchmarkevaluatingllms} versus deeper but domain-agnostic structures in JSONSchemaBench), and emphasis on format generation over context-based extraction.

	To address these limitations, we propose DeepJSONEval, a pioneering multilingual deep-nested JSON evaluation benchmark and framework. Our approach focuses on comprehension and extraction tasks, designed to comprehensively assess LLMs' ability to map raw text to given JSON schemas and return syntactically and semantically correct multi-layer nested JSON objects in web data mining contexts. The comparison of DeepJSONEval and other JSON-related benchmarks is illustrated in Table~\ref{tab:benchmark_comparison}.

	The workflow for constructing this benchmark dataset involves systematic data collection from diverse web sources, followed by schema conceptualization through hierarchical organization of extracted concepts, automated schema generation using our novel algorithm, and ground truth compilation from refined text corpora. Our Real-time Path-Value Updating Beam Exploration for Constrained Schema Subtree Construction algorithm ensures effective construction of complex nested structures required for comprehensive evaluation.

	Our approach introduces several innovations:
	\begin{itemize}
		\item An innovative \textbf{Real-time Path-Value Updating Beam Exploration for Constrained Schema Subtree Construction} is introduced that effectively supports the construction of complex nested structures.
		\item The evaluation framework is designed specifically targeting \textbf{deep nesting structures}, featuring JSON schemas with 3 to 7 levels of nesting depth with 17.5 properties in average, significantly enhancing evaluation complexity and real-world applicability in web data mining. Each field includes detailed descriptions to rigorously examine instruction-following capabilities of LLMs and precise information extraction under complex format constraints and semantic understanding requirements.
		\item \textbf{Comprehensive data type coverage} is implemented, incorporating strings, numbers, boolean values, string enumerations, and lists to systematically evaluate LLM robustness across different data types.
		\item A \textbf{multi-dimensional fine-grained evaluation} framework is developed encompassing format matching accuracy, field correctness, and complete structural correctness, providing multi-perspective and detailed assessment of JSON generation quality.
	\end{itemize}

	\begin{table*}[t]
	\centering
	\caption{Comparative Analysis of JSON Generation and Extraction Benchmarks: Multilingual Support, Task Design, Scale, Text Provided, Structural Complexity, Domain Label Provided, and Constrained Decoding Support (CDS).}
	\label{tab:benchmark_comparison}
	\resizebox{\textwidth}{!}{
		\begin{tabular}{lccccccc}
			\toprule
			\textbf{Benchmark} & \textbf{Multilingual} & \textbf{Task} & \textbf{Scale} & \textbf{Text} & \textbf{Depth} & \textbf{Label} & \textbf{CDS} \\
			\midrule
			StructEval \citep{yang2025structeval} & No & Structured Generation & 2035 (50 for JSON) & No & 2-4 & No & No \\
			json-mode-eval \cite{nousresearch_json-mode-eval}  & No & JSON Extraction & 100 & Yes & 1-2 & No & Yes \\
			STRUCTUREDRAG \cite{shorten2024structuredragjsonresponseformatting} & No & Format Following & 112 & Yes & 1-2 & No & No \\
			SoEval \cite{LIU2024103809} & Yes & JSON/XML Generation & 3700 (200 for JSON) & No & Not Specified & Yes & Yes \\
			FoFo \cite{xia2024fofobenchmarkevaluatellms} & No & Format Following & 493 & No & Not Specified & No & No \\
			NESTFULl \cite{basu2025nestfulbenchmarkevaluatingllms} & No & Function Calling  & 1800 & Yes & 1 & No & Yes \\
			StrucText-Eval \cite{gu2024structextevalevaluatinglargelanguage} & No & Structure Interpretation & 5800 & No & 1-3 & No & No \\
			JSONSchemaBench \cite{geng2025jsonschemabencharigorousbenchmarkof}  & No & JSON Generation & 6,000 & No & 2-23 & No & Yes \\
			Schema-Guided Dialogue \cite{rastogi2020scalablemultidomainconversationalagents} & No & Function Calling  & Not Specified & No & 1 & Yes & No \\
			\textbf{DeepJSONEval (Ours)} & \textbf{Yes} & \textbf{JSON Extractive} & \textbf{2,100} & \textbf{Yes} & \textbf{3-7} & \textbf{Yes} & \textbf{Yes} \\
			\bottomrule
		\end{tabular}
	}
\end{table*}

	DeepJSONEval establishes a new standard for objective and comprehensive evaluation of LLM structured output capabilities through its innovative evaluation dimensions, rigorous difficulty classification, detailed field descriptions, and large-scale multi-domain coverage. Our benchmark comprises 2100 high-quality data instances spanning ten diverse domains in web applications including Tourist Attraction Promotion, Electronic Devices Introduction, Patient Information, etc. We implement systematic difficulty grading based on nesting depth, categorizing 3-4 level structures as \emph{Medium} and 5-7 level structures as \emph{Hard}, providing progressive evaluation benchmarks for model capabilities. Through this framework, DeepJSONEval enables systematic assessment of LLMs' structured output capabilities and advances their reliability in real-world applications.

	\section{Related Work}
	Recent research has witnessed growing interest in evaluating and benchmarking LLMs across diverse capabilities and domains \cite{mitchener2025bixbenchcomprehensivebenchmarkllmbased}. RocketEval \cite{wei2025rocketevalefficientautomatedllmevaluation} demonstrates that lightweight LLMs can achieve comparable evaluation accuracy through structured checklist-based assessment, though it does not specifically address structured output validation. ThinkJSON \cite{agarwal2025thinkinsidethejsonreinforcement} tackled schema adherence through reinforcement learning, focusing primarily on model training rather than comprehensive evaluation.

	Critical limitations in existing benchmarks have been identified \cite{liu-etal-2025-empirical}, emphasizing the need for qualitative attributes such as diversity, redundancy, and difficulty assessment. Data contamination significantly impacts evaluation validity, particularly in larger models \cite{kocyigit2025overestimationinllmevaluationa}. Domain-specific evaluation frameworks like IberoBench \cite{baucells-etal-2025-iberobench} and Evalita-LLM \cite{magnini2025evalitallmbenchmarkinglargelanguagemodels} have developed comprehensive multilingual benchmarks, while Language Ranker \cite{Li_Shi_Liu_Yang_Payani_Liu_Du_2025} proposed metrics for quantifying LLM performance across diverse languages.

	Automated evaluation approaches have emerged to address scalability challenges. CodeArena \cite{du2025codearenaacollectiveevaluationplatform} introduced collective evaluation mechanisms for code generation, while annotation costs have been reduced by combining human and synthetic feedback \cite{zhou2025acceleratingunbiasedllmevaluationvia}. However, systematic JSON evaluation benchmarks remain absent, creating a gap in assessing structured output capabilities essential for web data mining applications.

	\begin{figure*}[t]
		\centering
		\includegraphics[width=0.8\textwidth]{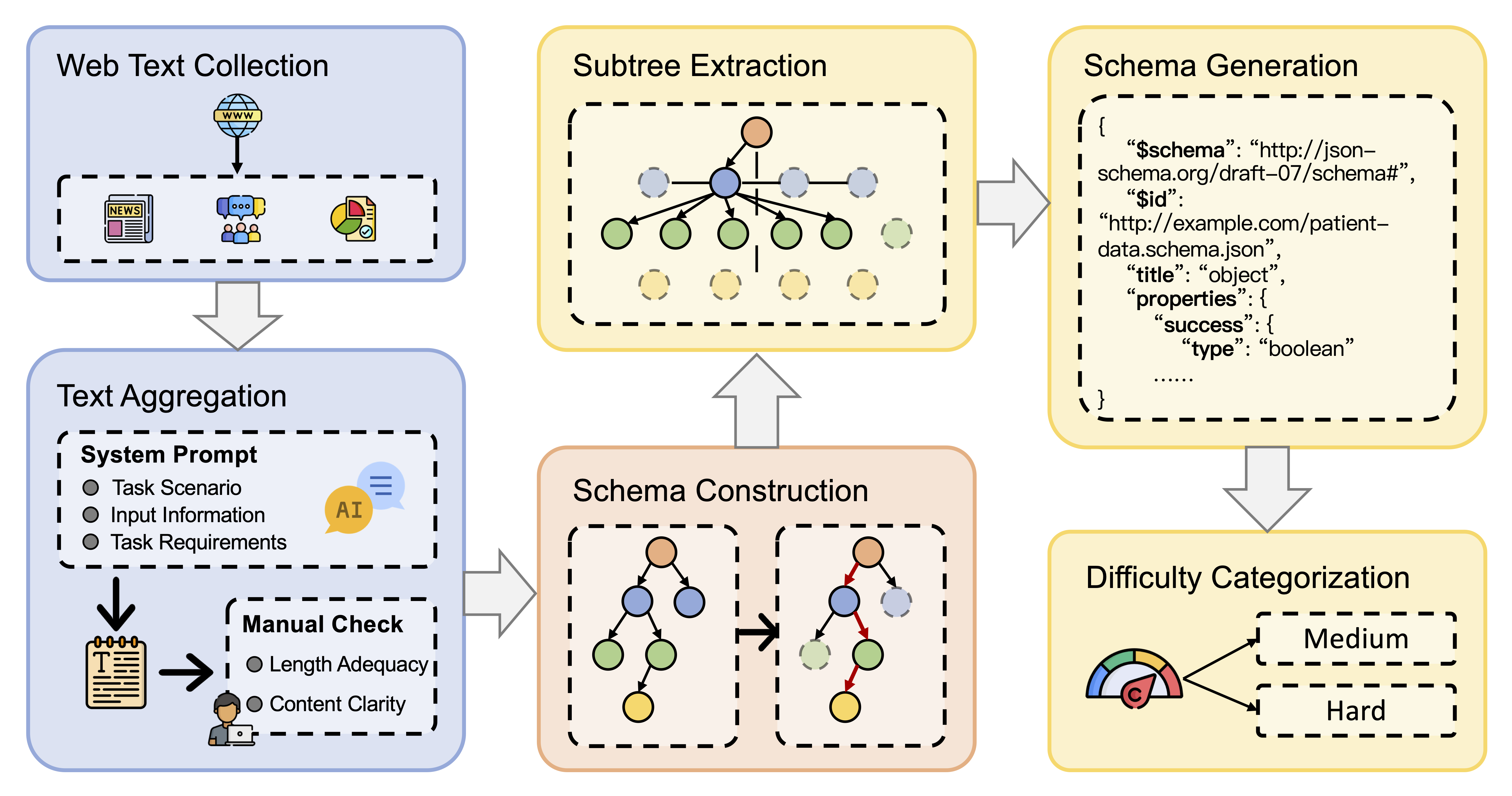}
		\caption{The workflow of benchmark construction.}
		\label{method_flow}
	\end{figure*}

	\section{Method}
	\label{method}
	
	The construction of DeepJSONEval follows a systematic four-stage workflow designed to generate high-quality, domain-diverse evaluation instances with complex nested structures. Figure~\ref{method_flow} illustrates our comprehensive pipeline, which transforms raw web content into a rigorous benchmark through automated processing and careful validation.

	\subsection{Web Text Collection and Multi-Document Aggregation}
	\label{text_collection}
	
	To address the information dispersion and redundancy inherent in web corpora across heterogeneous sources, we implement a systematic multi-text aggregation and rewriting strategy that synthesizes complementary content while eliminating cross-source redundancy. We employ LLMs to execute multi-text aggregation and rewriting by processing raw web text using the following prompt template:
	
	\begin{framed}
		\noindent\textbf{Role} \newline
		You are an expert editor specializing in multi-document synthesis. Read multiple input texts and produce one concise, information-dense summary that merges overlapping content, resolves contradictions when possible, and preserves key facts.\newline
		
		\noindent\textbf{Guide for Inputs} \newline
		DOCS: [A list/array of documents (plain text). Each item may be a paragraph, article, or note.] \newline
		
		\noindent\textbf{Rules} \newline
		1. Aggregation: Merge redundant points; group related ideas; eliminate repetition. \newline
		2. Compression: Prefer shorter phrasing and high signal-to-noise ratio; avoid filler, hedging, and rhetorical questions. \newline
		3. Faithfulness: Do not invent facts. Only use information present in the inputs. If sources disagree, briefly note the disagreement. \newline
		4. Clarity \& Flow: Use precise vocabulary, active voice, and logical order (problem → evidence → implications or theme → key points → takeaway). \newline
		5. Style: Neutral, objective, and professional. Avoid bullet lists unless the content clearly benefits from this format. \newline
		6. Language: Use the same language as the source DOCS for output. \newline
		7. Length: Generate at least 1500 words. \newline
		
		\noindent\textbf{DOCS} \newline
		[INPUT DOCUMENTS] \newline
	\end{framed}
	
	Through this prompt template, we instruct the LLM to integrate complementary aspects (methods, datasets, effect sizes) that rarely co-occur in single sources while removing inter-source redundancy and prioritizing data-bearing propositions. This approach enables the synthesis of coherent, comprehensive summaries that preserve factual accuracy while achieving significant compression ratios.
	
	After the text-schema pair constructed, we apply a lightweight human-in-the-loop protocol (seed Apendix~\ref{hqc}) to (i) verify semantic faithfulness of LLM-constructed items, (ii) reduce teacher bias/leakage risk, and (iii) release a reliable \textit{Gold} set.
	
	\subsection{Schema Tree Construction}
	\label{schema_tree}
	
	Following text aggregation, we systematically identify and extract key terms and concepts from the collected corpus. These elements are organized into hierarchical tree structures where nodes represent distinct properties and edges capture inheritance relationships between parent and child properties (e.g., \textit{author} $\rightarrow$ \textit{author.name} $\rightarrow$ \textit{author.name.first}).
	
	The schema tree construction process leverages domain expertise and linguistic analysis to ensure that the resulting hierarchical structures accurately reflect the semantic relationships inherent in the source content. Each tree serves as a comprehensive representation of the information space within a specific domain, providing the foundational structure for subsequent subtree extraction and schema generation.
	
	\subsection{Real-time Path-Value Updating Beam Search for Constrained Schema Subtree Construction}
	\label{algorithm_intro}
	
	\begin{algorithm}[ht]
		\caption{Real-time Path-Value Updating Exploration for Subtree Extraction}
		\label{alg:rpu}
		\begin{algorithmic}[1]
			\Require Tree $\mathcal{T}=(V,E)$; $d_{\min},d_{\max}$; $n_{\min},n_{\max}$; $\texttt{top\_k}$
			\Ensure A set of subtrees $\mathcal{S}$
			\State $\mathcal{S}\gets \varnothing$
			\For{root $r\in V$}
			\State $S\gets (\{r\},\varnothing)$
			\While{$d(S)<d_{\max}$ \textbf{and} $|V_S|<n_{\max}$}
			\State Generate candidate path set $\mathcal{P}(S)$ with simple paths
			starting at $\mathcal{F}(S)$ and length $\le L_{\max}$
			where $L_{\max}\!=\!\min\{d_{\max}-d(S),\ n_{\max}-|V_S|\}$
			\ForAll{$p\in\mathcal{P}(S)$}
			\State Compute $\mathrm{Val}(p\mid S)$ via \eqref{eq:val}
			\EndFor
			\State $\mathcal{B} \gets \text{TopK}\bigl(\mathcal{P}(S), \mathrm{Val}, \texttt{top\_k}\bigr)$
			\ForAll{$p\in \mathcal{B}$}
			\If{$\mathrm{Feas}(S\oplus p)=1$}
			\State $S\gets S\oplus p$
			\EndIf
			\EndFor
			\If{$\mathcal{B}=\varnothing$}
			\State \textbf{break}
			\EndIf
			\EndWhile
			\If{$d(S)\ge d_{\min}$ \textbf{and} $|V_S|\ge n_{\min}$}
			\State $\mathcal{S}\gets \mathcal{S}\cup\{S\}$
			\EndIf
			\EndFor
			\State \Return $\mathcal{S}$
		\end{algorithmic}
	\end{algorithm}
	
	\begin{table*}[t]
		\centering
		\caption{The leaderboard of DeepJSONEval with leading LLMs}
		\label{tab:data}
		\begin{tabular}{lrrrrrrrrr}
			\toprule
			\multirow{2}{*}{Model} & \multicolumn{3}{c}{Overall} & \multicolumn{3}{c}{Medium} & \multicolumn{3}{c}{Hard} \\
			\cmidrule(lr){2-4} \cmidrule(lr){5-7} \cmidrule(lr){8-10}
			& Syntax & Key & Strict & Syntax & Key & Strict & Syntax & Key & Strict \\
			\midrule
			Claude Sonnet 4 & \textbf{99.05} & \textbf{90.73} & 57.90 & \textbf{100.00} & 94.52 & 69.51 & \textbf{98.61} & \textbf{89.01} & 52.63 \\
			Magistral Medium 2506 & 98.10 & 90.36 & \textbf{59.81} & \textbf{100.00} & \textbf{95.50} & \textbf{71.34} & 97.22 & 88.04 & \textbf{54.57} \\
			DeepSeek R1 0528 & 97.90 & 89.57 & 57.33 & \textbf{100.00} & 94.56 & 68.29 & 96.95 & 87.30 & 52.35 \\
			Gemini 2.5 Pro & 97.52 & 89.00 & 56.19 & \textbf{100.00} & 94.78 & 70.73 & 96.40 & 86.37 & 49.58 \\
			Qwen3 235B A22B & 97.14 & 88.33 & 56.19 & \textbf{100.00} & 94.95 & 67.07 & 95.84 & 85.33 & 51.25 \\
			DeepSeek R1 Distill Llama 70B & 95.44 & 88.15 & 58.29 & 99.39 & 93.80 & 67.68 & 93.63 & 85.58 & 54.02 \\
			Magistral Small 2506 & 93.71 & 85.75 & 53.52 & 99.39 & 93.46 & 66.46 & 91.14 & 82.25 & 47.65 \\
			Qwen3 30B A3B & 93.71 & 84.66 & 52.76 & \textbf{100.00} & 93.30 & 66.46 & 90.86 & 80.74 & 46.54 \\
			Llama 4 Maverick & 92.19 & 84.17 & 54.48 & 95.73 & 89.69 & 64.63 & 90.58 & 81.67 & 49.86 \\
			Qwen3 14B & 91.81 & 83.39 & 51.24 & 96.95 & 91.04 & 62.20 & 89.47 & 79.92 & 46.26 \\
			Hunyuan A13B & 93.33 & 83.11 & 48.38 & 98.17 & 89.96 & 54.88 & 91.14 & 79.99 & 45.43 \\
			Qwen3 32B & 87.81 & 80.18 & 50.67 & 96.95 & 91.87 & 68.29 & 83.66 & 74.87 & 42.66 \\
			\bottomrule
		\end{tabular}
	\end{table*}
	
	\paragraph{\textbf{Algorithm Overview}}
	Given a rooted property tree $\mathcal{T}=(V,E)$, where $V$ represents properties and $E$ contains directed edges from parent properties to their children (e.g., \textit{author $\rightarrow$ author.name}), we aim to extract a set of subtrees $\mathcal{S}=\{S_1,S_2,\ldots\}$ by iteratively expanding paths from the current subtree frontier while updating path values in real time. Each expansion must satisfy constraints on schema depth and the number of properties.
	
	\textbf{Algorithm Inputs:}
	\begin{itemize}
		\item Tree $\mathcal{T}=(V,E)$ with $|V|=n$ nodes.
		\item Depth bounds: minimum target schema depth $d_{\min}$ and maximum target schema depth $d_{\max}$.
		\item Size bounds: minimum number of properties $n_{\min}$ and maximum number of properties $n_{\max}$.
		\item Beam width $\texttt{top\_k}\in\mathbb{N}^+$: the maximum number of candidate paths to retain per expansion round.
	\end{itemize}
	
	\textbf{Algorithm Outputs:} A collection of high-value subtrees $\mathcal{S}$ such that every $S\in\mathcal{S}$ satisfies $d(S)\!\in\![d_{\min},d_{\max}]$ and $|V_S|\!\in\![n_{\min},n_{\max}]$, where $d(S)$ denotes the depth of $S$.
	
	\paragraph{\textbf{Real-time Path-Value Updating Mechanism}}
	As shown in Algorithm \ref{alg:rpu}, let the current subtree be $S=(V_S,E_S)$ and define the frontier $\mathcal{F}(S)=\{u\in V\setminus V_S \mid \exists v\in V_S,\ (v,u)\in E\}$. A \emph{candidate path} is $p=(u_0,\ldots,u_\ell)$, where $u_0\in\mathcal{F}(S)$, each $(u_{i-1},u_i)\in E$, and all $u_i\notin V_S$. We denote by $S\oplus p$ the subtree after augmenting $S$ with all nodes and edges of $p$.
	
	Given an association score function $\mathrm{Assoc}:V\times V\to[0,1]$ (see Appendix), the correlation of a node $u$ to the current subtree is defined as:
	\begin{equation}
		\mathrm{Corr}(u\mid S) \;=\; \max_{v\in V_S}\ \mathrm{Assoc}(u,v).
		\label{eq:corr}
	\end{equation}
	For a new node $u\notin V_S$, we define its marginal contribution as:
	\begin{equation}
		\Delta(u\mid S) \;=\; \alpha\cdot \mathrm{Corr}(u\mid S),
		\quad \alpha>0.
		\label{eq:delta}
	\end{equation}
	
	Let $d(S)$ denote the current depth and $|V_S|$ the current size. After augmenting with path $p$, we incorporate \emph{soft window} rewards that peak within the target intervals:
	\begin{align}
		R_{\text{depth}}(S\oplus p)
		&= 1 - \frac{\left|\,\mathrm{clip}\!\bigl(d(S\oplus p),\,d_{\min},d_{\max}\bigr) - d(S\oplus p)\,\right|}
		{d_{\max}-d_{\min}+\varepsilon}, \label{eq:rdepth}\\
		R_{\text{size}}(S\oplus p)
		&= 1 - \frac{\left|\,\mathrm{clip}\!\bigl(|V_{S\oplus p}|,\,n_{\min},n_{\max}\bigr) - |V_{S\oplus p}|\,\right|}
		{n_{\max}-n_{\min}+\varepsilon},
		\label{eq:rsize}
	\end{align}
	where $\mathrm{clip}(x,a,b)=\min\{\max\{x,a\},b\}$ and $\varepsilon>0$ prevents division by zero.
	
	To ensure structural validity, we penalize infeasible configurations:
	\begin{equation}
		\mathrm{Penalty}(p\mid S)\;=\; 
		\mathbb{I}[u_0\notin \mathcal{F}(S)] +
		\sum_{i=1}^{\ell}\mathbb{I}\bigl[(u_{i-1},u_i)\notin E\bigr].
		\label{eq:penalty}
	\end{equation}
	Both rewards achieve a maximum value of 1 within their respective target intervals and decrease linearly outside these bounds.
	
	Let $\gamma\in(0,1]$ be a discount factor for deeper steps along the path. For $p=(u_0,\ldots,u_\ell)$, we define the path-value function as:
	
	\begin{equation}
		\begin{aligned}
			\mathrm{Val}(p\mid S)&=
			\sum_{i=0}^{\ell}\gamma^{i}\,\Delta\!\left(u_i\ \middle|\ S\cup\{u_0,\ldots,u_{i-1}\}\right) \\
			&+\lambda_d\,R_{\text{depth}}(S\oplus p)
			+\lambda_n\,R_{\text{size}}(S\oplus p) \\
			&-\eta\,\mathrm{Penalty}(p\mid S)
		\end{aligned}
		\label{eq:val}
	\end{equation}
	with non-negative weights $\lambda_d,\lambda_n,\eta\ge 0$.
	
	To reject infeasible augmentations, we define the feasibility function:
	\begin{equation}
		\mathrm{Feas}(S\oplus p)=
		\begin{cases}
			0, & \text{if } d(S\oplus p)>d_{\max}\ \text{or}\ |V_{S\oplus p}|>n_{\max},\\
			1, & \text{otherwise.}
		\end{cases}
		\label{eq:feas}
	\end{equation}
	When $\mathrm{Feas}(S\oplus p)=0$, we set $\mathrm{Val}(p\mid S)=-\infty$ and discard path $p$.
	
	In practice, we expand subtrees by iteratively generating candidate paths, updating their values using Equation \eqref{eq:val}, and selecting up to $\texttt{top\_k}$ best paths to augment the current subtree.
	
	\subsection{Schema Generation and Benchmark Ground Truth Construction}
	\label{ground_truth}
	
	We systematically convert extracted subtrees into formal JSON schemas with detailed specifications and implement difficulty categorization based on nesting depth: \emph{Medium} (3-4 levels) and \emph{Hard} (5-7 levels). Domain experts annotate ground truth data through systematic mapping and validation, with human-in-the-loop quality control (see Appendix) ensuring semantic faithfulness and establishing a reliable \textit{Gold} standard.
	
	\section{Data Statistics}
\begin{figure}[t]
		\centering
		\includegraphics[width=0.45\textwidth]{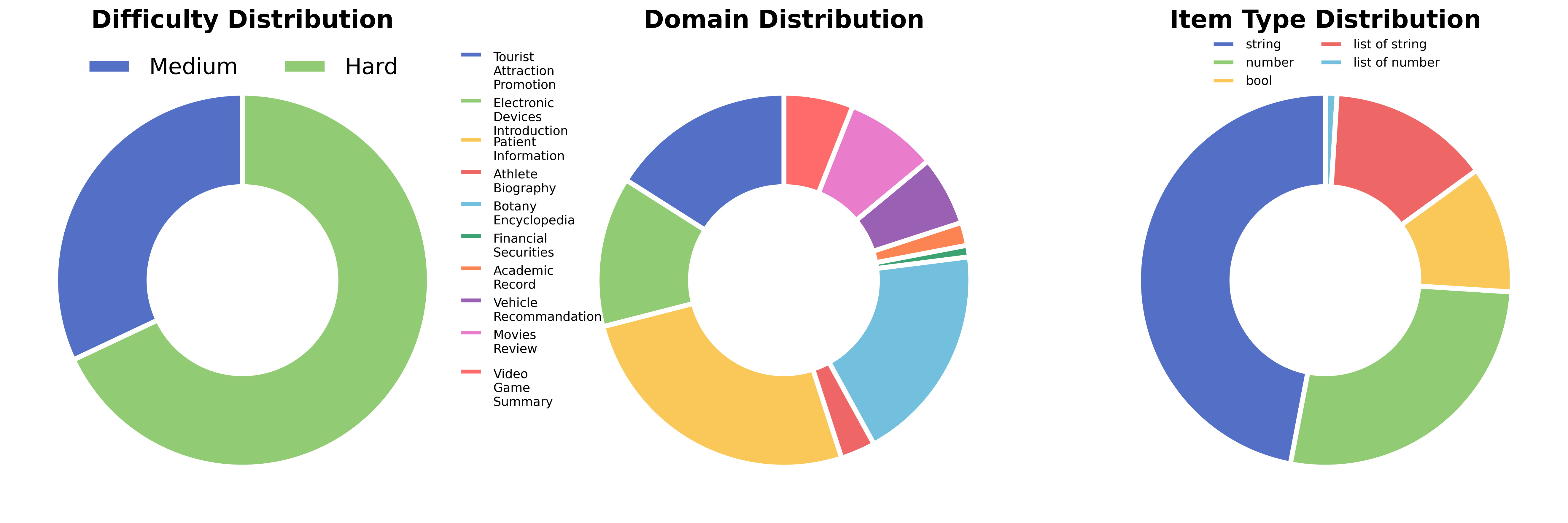}
		\caption{The distribution overview of DeepJSONEval across difficulty levels, domains, and categories.}
		\label{overview}
	\end{figure}

	Using this methodology, we construct DeepJSONEval comprising 2100 data instances across ten web domains: Tourist Attraction Promotion, Electronic Devices Introduction, Patient Information, Athlete Biography, Botany Encyclopedia, Financial Securities, Academic Record, Vehicle Recommendation, Movie Review, and Video Game Summary. Instances are categorized by structural complexity: \emph{Medium} difficulty (3-4 nesting levels) and \emph{Hard} difficulty (5-7 nesting levels), with comprehensive distributions shown in Figure \ref{overview}.
	
\begin{figure}[t]
		\centering
		\includegraphics[width=0.45\textwidth]{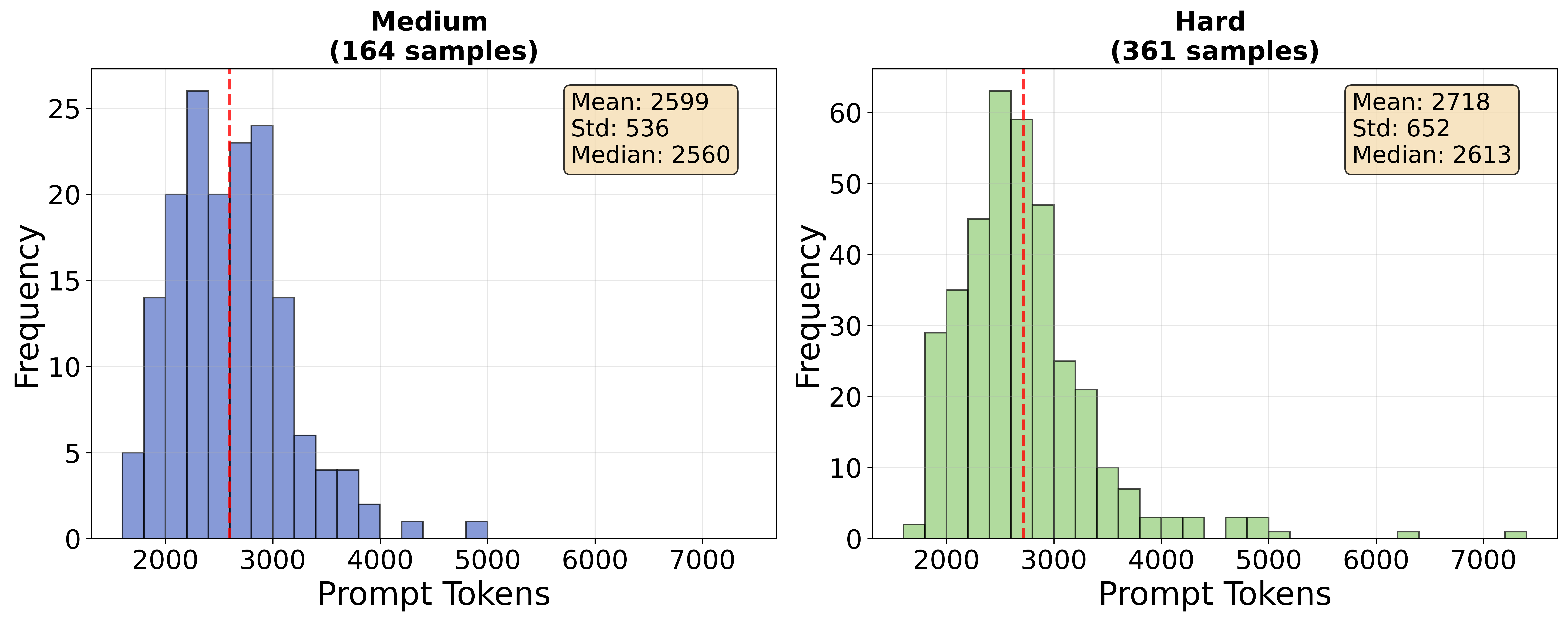}
		\caption{The prompt length statistics for Medium and Hard samples in DeepJSONEval.}
		\label{prompt_length}
	\end{figure}

	The prompt length distribution demonstrates a reasonable progression in complexity, with Hard samples exhibiting longer average lengths and greater variability, which appropriately reflects the increased structural complexity associated with deeper JSON nesting levels (see Figure~\ref{prompt_length}). The concentrated distribution around 2000-4000 tokens provides an optimal range for evaluating JSON parsing capabilities without introducing excessive computational overhead, while the substantial sample sizes (164 Medium, 361 Hard) ensure robust statistical evaluation across difficulty tiers.
	
	\begin{figure*}[t]
		\centering
		\includegraphics[width=1.0\textwidth]{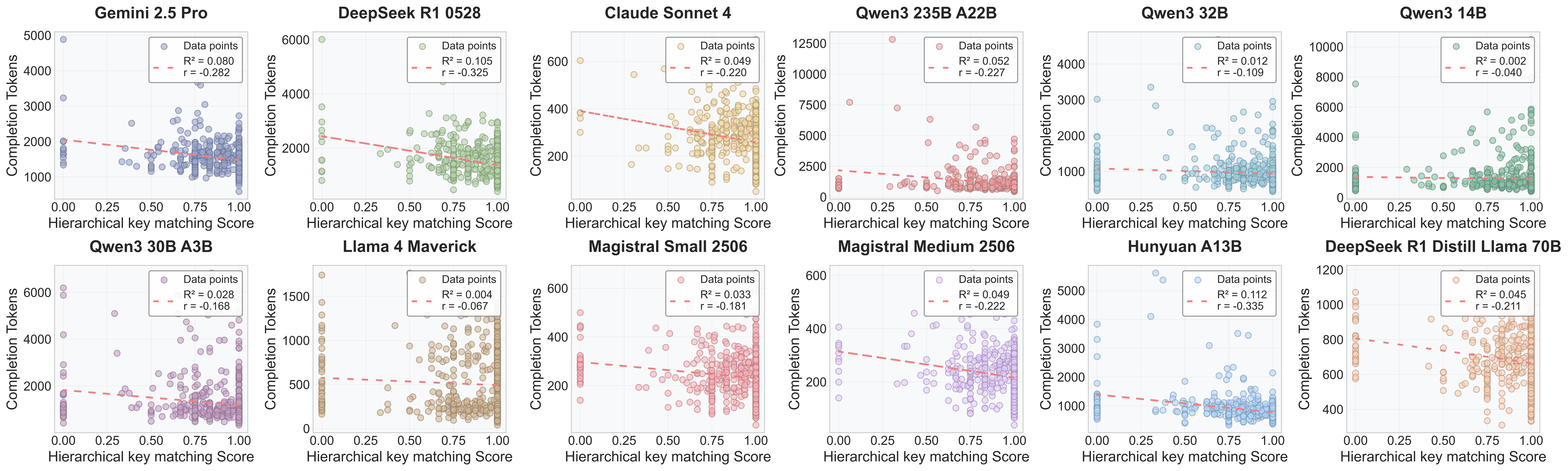}
		\caption{The response length distribution across evaluation scores for multiple LLMs.}
		\label{length_response}
	\end{figure*}
	
	\begin{figure*}[t]
		\centering
		\includegraphics[width=1.0\textwidth]{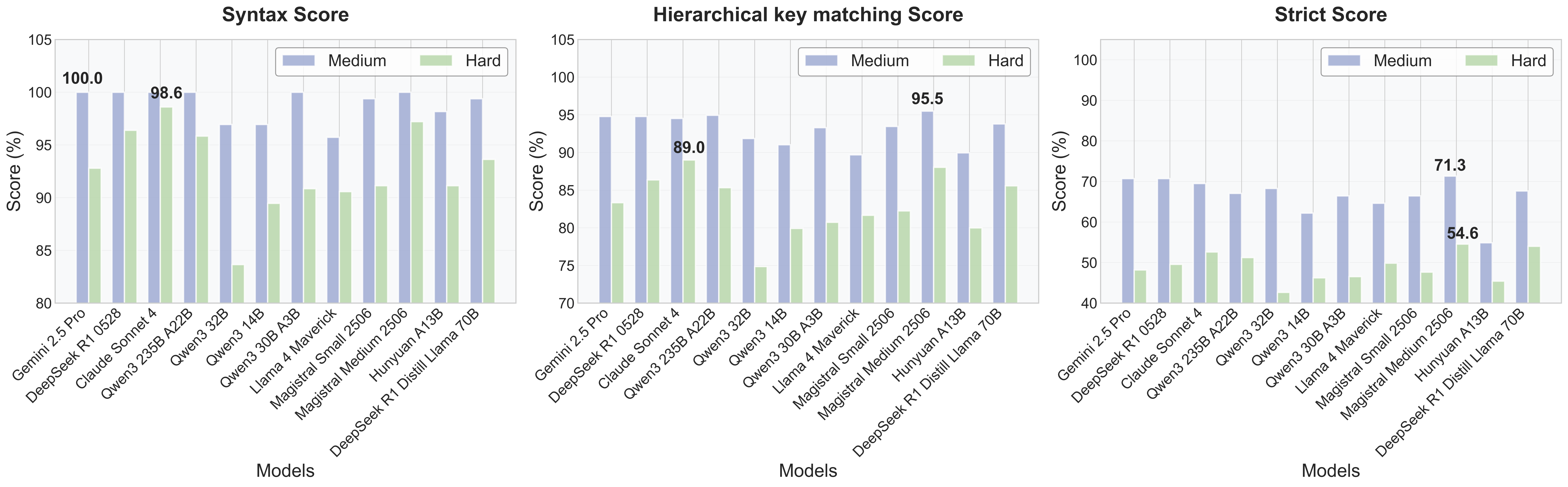}
		\caption{The performance comparison across format, detailed, and strict evaluation metrics for Medium and Hard difficulty levels.}
		\label{diff_level}
	\end{figure*}
	
	\section{Experiments}
	
	\label{experiment}
	
	\subsection{Experimental Setup}
	
	The benchmark generation experiments were conducted using two distinct computational environments optimized for different generation tasks. The text aggregation task utilized the \texttt{DeepSeek-R1} model running on \texttt{Ascend 910B2} processors, configured with an inference temperature of 1 to promote diverse and creative text rewriting. This dual-environment setup was designed to leverage the specific computational strengths of each hardware platform while optimizing model performance for distinct generation requirements. The evaluation metrics are detailed in Appendix. 
	
	\subsection{Evaluation on Leading LLMs}
	\paragraph{\textbf{Overall.}}
	
	Table~\ref{tab:data} presents the performance of 12 representative LLMs with leading capabilities, selected from the OpenCompass LLM Leaderboard \cite{opencompassleaderboard}, on the DeepJSONEval benchmark. The models are ranked in descending order by their overall detailed scores. The detailed score measures the comprehensive extraction capabilities of models, with higher scores indicating stronger extraction performance. However, in specific agentic systems where complete accuracy is required, the strict score provides a more accurate reflection of model capabilities.

	\paragraph{\textbf{Response Length Analysis}}
	In Figure~\ref{length_response}, analysis of response length versus evaluation scores across 12 LLMs reveals minimal correlations ($r$ ranging from -0.335 to -0.040), with most models showing $R^2 < 0.05$, indicating that response length accounts for less than 5\% of score variation. This demonstrates that DeepJSONEval effectively evaluates structural accuracy and semantic correctness rather than verbosity, validating our multi-dimensional evaluation framework's ability to assess genuine JSON generation capabilities independent of output length.
	
	\paragraph{\textbf{Performance versus difficulties}}
	Performance analysis across medium (3-4 levels) and hard (5-7 levels) difficulty tiers reveals systematic degradation in all LLMs, as shown in Figure~\ref{diff_level}, with strict evaluation scores exhibiting the most substantial declines (17.22\%-37.53\%) compared to format scores (1.39\%-13.71\%). Hard-level tasks consistently challenge all models, with strict scores remaining below 60\%, demonstrating that deep nesting structures effectively differentiate model capabilities. These findings validate DeepJSONEval's discriminative power and difficulty stratification, establishing a robust framework for evaluating structured output generation across varying complexity levels.
	
\begin{figure*}[t]
		\centering
		\includegraphics[width=1.0\textwidth]{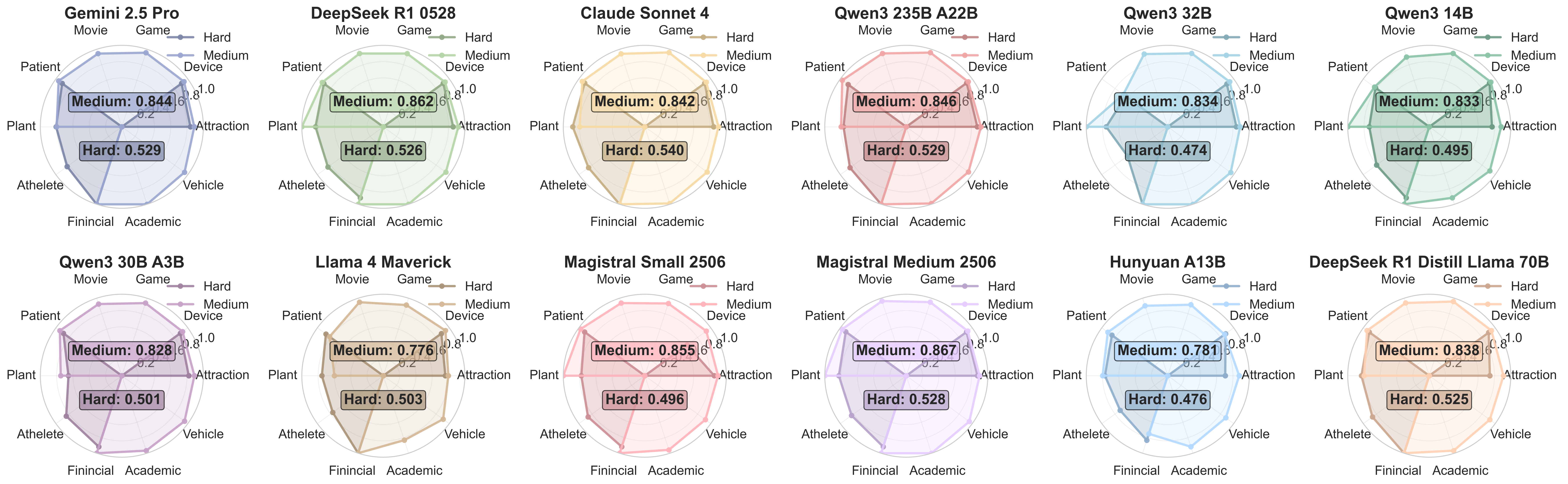}
		\caption{The Hierarchical key matching Score across domains and difficulty levels for multiple LLMs.}
		\label{radar}
	\end{figure*}

	\paragraph{\textbf{Performance in diffrent domains}}
	Cross-domain performance analysis reveals consistent model behavior across 10 domains, with medium difficulty scores ranging 0.776-0.867 and hard difficulty scores spanning 0.474-0.540, indicating minimal domain-specific variance (see Figure~\ref{radar}). The selective distribution of hard difficulty samples across domains reflects realistic JSON application scenarios where complex nesting requirements are domain-dependent. This balanced performance distribution validates DeepJSONEval's ecological validity and confirms that the benchmark accurately represents practical deployment challenges.
	
\begin{figure}[t]
		\centering
		\includegraphics[width=0.45\textwidth]{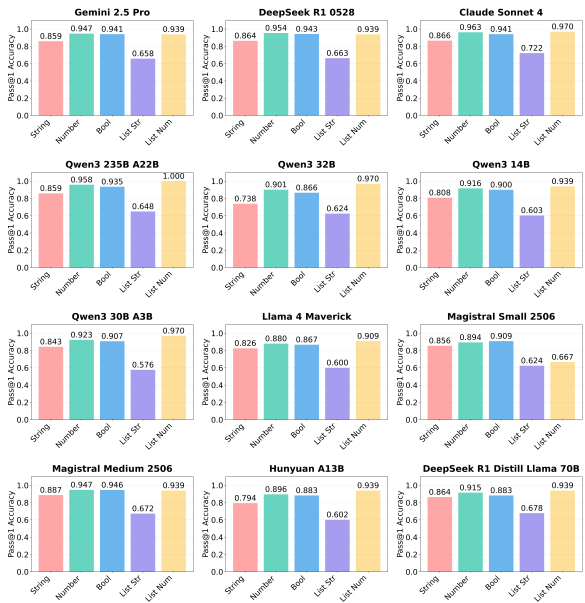}
		\caption{The performance comparison across JSON element types for multiple LLMs.}
		\label{item_type}
	\end{figure}

	\paragraph{\textbf{Performance in different data types}}
	Figure~\ref{item_type} demonstrates systematic performance variation across JSON element types, with LLMs achieving highest accuracy on numeric lists (0.90-1.00) and lowest on string lists (0.576-0.722). This consistent hierarchy across all 12 models reveals fundamental limitations in processing complex nested list structures compared to primitive data types. The pronounced performance degradation on hierarchical list elements indicates architectural challenges in structured output generation, highlighting the need for improved training strategies targeting nested JSON relationships.
	
	\paragraph{\textbf{Qualitative Analysis}}
	see Appendix.

	\section{External Validity via a Small End-to-End Web Pipeline}
	\label{sec:external-validity}
	
	We examine whether performance on \textsc{DeepJSONEval} predicts real-world utility in a practical Web pipeline:
	\emph{crawl $\rightarrow$ extract $\rightarrow$ JSON $\rightarrow$ query/analytics}.
	We quantify the association between benchmark scores and real world pipeline scores.
	
	\subsection{Experiment Setting}
	
	\subsubsection{Pipeline} 
	\begin{enumerate}
		\item \textbf{Crawl}: domain-allowlisted seed URL.
		\item \textbf{Preprocess}: boilerplate removal, language detection, schema construction.
		\item \textbf{Extract}: run candidate LLMs to produce schema-conformant JSON per document.
		\item \textbf{Query/Analytics}: calculate the correlation of \textbf{DeepJSONEval} score and pipeline scores.
	\end{enumerate}
	
	\subsubsection{Evaluation Criteria}
	See Appendix.	
	
	\subsubsection{Models \& Domains Selection}
	We select 3 LLMs that represent a wide range of \textbf{DeepJSONEval} scores (low$\to$high): Qwen3 32B, DeepSeek R1 Distill Llama 70B and Claude Sonnet 4 with 3 domains: Athlete Biography, Vehicle Recommendation and Video Game Summary, each domain containing 2 data units with hard-level schemas.
	
	\subsection{Experiment Result \& Analysis}
	
	\begin{table}[t]
		\centering
		\caption{Result of End-to-End Pipeline}
		\label{end2end}
		\begin{tabular}{lc}
			\toprule
			Model & key matching score \\
			\midrule
			Claude Sonnet 4  & 97.36 \\
			DeepSeek R1 Distill Llama 70B  & 91.87 \\
			Qwen3 32B  & 83.26 \\
			\bottomrule
		\end{tabular}
	\end{table}
	
	The correlation between 3 model key matching scores in Table~\ref{end2end} and those in Table~\ref{tab:data} is $0.987$. This provides external validity: \textbf{DeepJSONEval} gains predict concrete improvements in end-to-end Web extraction and analytics. We also note that the models in the end-to-end pipeline achieve higher scores than those in the benchmark. This discrepancy arises because the evaluation set consists of aggregated original texts, resulting in high information density that may cause models to confuse or overlook information. In contrast, the pipeline employs raw web pages with lower information density, which facilitates more accurate information extraction by the models. 
	
	\section{Conclusion}
	
	This work addresses the critical limitations in evaluating multi-layer nested JSON generation and information extraction capabilities of LLMs by introducing \textbf{DeepJSONEval}, a pioneering benchmark framework for deep nested JSON structure evaluation. Our innovative tree-based schema generation algorithm successfully constructs complex nested structures ranging from 3 to 7 layers deep, establishing the first graded difficulty assessment framework that transitions from pseudo schema to standard schema with exceptional scalability. It enables extensible synthesis of domain-specific question-answer pairs with customizable nesting depths, allowing for the generation of standardized JSON structured extraction datasets across diverse application scenarios. The comprehensive evaluation framework achieves multi-dimensional assessment through format matching, field correctness, and structural integrity metrics, systematically evaluating model performance across 2100 high-quality instances spanning 10 diverse domains including tourism, healthcare, entertainment, etc. Systematic difficulty grading of DeepJSONEval distinguishes medium complexity from hard scenarios, providing fine-grained differentiation suitable for current leading models and emerging development trends. Future work will focus on extending the framework to support even deeper nesting levels and incorporating dynamic schema adaptation to further enhance the applicability of benchmark to evolving LLM architectures.

	\section*{Future Work}
	\subsection*{Time Complexity Optimization of Algorithm} 
	
	In Section \ref{algorithm_intro}, we outline algorithmic directions that explicitly target the dominant factors in the current complexity $T_{\text{total}} = O\!\big(R \cdot L \cdot b^{L}\big)$, where $b$ is the (effective) branching factor, $L$ is the candidate path length cap, and $R$ is the number of expansion rounds. The optimization process will focus on the following aspects:
	
	\begin{itemize}
		\item \textbf{Lazy Valuation with Prefix Reuse}: maintain prefix accumulators $\sum_{i\le\ell} \gamma^{i}\Delta(u_i\mid\cdot)$ and an incremental cache for $\max_{v\in V_S}\!\mathrm{sim}(u,v)$; evaluate window rewards only at interval crossing checkpoints
		\item \textbf{Tree-DP for Pure Trees}: bottom-up DP to precompute, for each node, top-$t$ outward path scores for lengths $1..L$ (with discount and cached correlations)
		\item \textbf{Parallel Valuation and Top-k}: SIMD/GPU batching for correlations and rewards; thread-local heaps with periodic $k$-way merges; 64-bit hashing for deduplication
	\end{itemize}
	
	These optimization strategies target the exponential growth in candidate path evaluation by implementing prefix reuse and dynamic programming techniques, reducing the effective branching factor from $b^L$ to approximately $b\log{L}$ for practical tree structures. The parallel processing approach enables efficient utilization of multi-core architectures, achieving near-linear speedup for correlation computations across independent subtree branches.
	
	\bibliography{custom}
	
	\newpage
	\appendix
	\section{Evaluation Criteria}
	\label{criteria}
	
	\begin{figure}[htbp]
		\centering
		\includegraphics[width=0.46\textwidth]{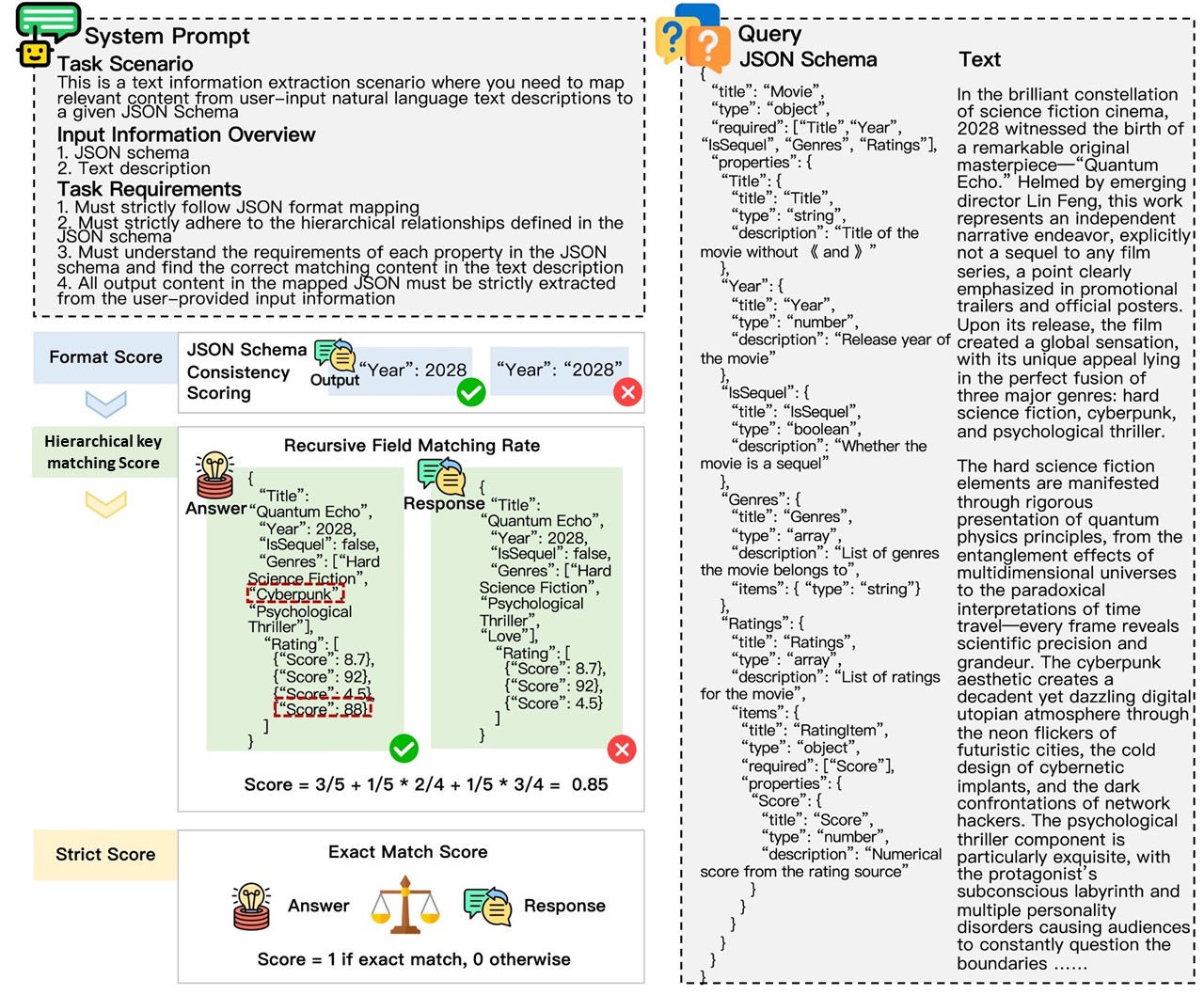}
		\caption{The illustrative example of the DeepJSONEval multi-dimensional assessment criteria framework.}
		\label{criterion}
	\end{figure}
	
	Upon completion of the benchmark dataset construction, we establish a comprehensive evaluation framework, which is displayed in Figure \ref{criterion}, comprising four distinct criteria to systematically assess LLM performance on our synthesized benchmark. These evaluation metrics are designed to provide a multi-dimensional analysis of model capabilities across different aspects of the structured data generation and comprehension tasks.
	
	\begin{itemize}
		\item \textbf{Criterion 1: Syntax Score}
		\newline This criterion evaluates LLMs' ability to generate syntactically valid JSON outputs through sequential validation of parsing and schema conformance, with scoring defined as:
		\begin{equation}
			S_{syntax} = \mathds{1}[\mathrm{valid}(\mathrm{output}) \land \mathrm{match}(\mathrm{output}, \mathrm{schema})]
		\end{equation}
		
		\item \textbf{Criterion 2: Hierarchical key matching Score}
		\newline Based on the \textit{key\_mathcing\_score} in JSON-Based Reward of ThinkJSON \cite{agarwal2025thinkinsidethejsonreinforcement}, considering the multi-layer nested structrue, this criterion performs comprehensive property-wise comparison with uniform weighting across all hierarchical levels based on Jaccard similarity, computing weighted differences through systematic structural traversal:
		\begin{equation}
			S_{key} = \frac{|p_{out} \cap p_{truth}|}{|p_{out} \cup p_{truth}|}
		\end{equation}
		
		\item \textbf{Criterion 3: Strict Score}
		\newline This criterion implements binary exact-match evaluation through strict equality verification between LLM output and ground truth JSON:
		\begin{equation}
			S_{strict} = \mathds{1}[\mathrm{output} \equiv \mathrm{truth}]
		\end{equation}
		
	\end{itemize}
	
	\section{Node Association Score}
	\label{node_association}
	\textbf{Inputs}: The discription of node $u$ and node $v$, $\mathrm{desc}_u$ \& $\mathrm{desc}_v$
	
	\noindent\textbf{Outputs}: the \emph{association} score $\mathrm{Assoc}(u,v)\in[0,1]$ between 2 nodes $u$ and $v$
	
	Let $d(u,v)\in\{1,2,\dots\}$ be the shortest-path distance between $u$ and $v$ in the property node tree, the penalty of distance is
	\begin{equation}
		\kappa_{\text{dist}}(u,v)=\exp\!\big(-\alpha\, d(u,v))
	\end{equation}
	
	The association of node description is:
	\begin{equation}
		s_{\text{desc}}(u,v)=\frac{\langle \phi(\mathrm{desc}_u),\,\phi(\mathrm{desc}_v)\rangle}{\|\phi(\mathrm{desc}_u)\|\,\|\phi(\mathrm{desc}_v)\|}
	\end{equation}
	where $\phi(\cdot)$ is normalized text embedding.
	
	The Association function is finally:
	\begin{equation}
		\mathrm{Assoc}(u,v)=\kappa_{\text{dist}}(u,v)\cdot\Big(\lambda_{\text{name}}\, s_{\text{name}}(u,v)+\lambda_{\text{desc}}\, s_{\text{desc}}(u,v)\Big)
	\end{equation}
	with $\lambda_{\text{name}},\lambda_{\text{desc}}\ge 0$, $\lambda_{\text{name}}+\lambda_{\text{desc}}=1$.
	
	\section{Qualitative Analysis Examples}
	\label{qae}
	This section presents two representative examples across different difficulty levels. Comparative analysis reveals that increased nesting depth in hard-level tasks correlates with higher error prevalence and diversity compared to medium-level tasks. The hard-level case exhibits multiple error categories: type mismatches (e.g., returning string values such as "4 GB" instead of required numerical data), semantic hallucinations (e.g., property repetition or introduction of non-existent entities like "personal wellness hub"), and extraction inaccuracies (e.g., incomplete proper noun extraction). Conversely, the medium-level case demonstrates only isolated type errors. These findings substantiate the hypothesis that increased nesting complexity introduces systematic challenges including field omissions, hierarchical misalignments, type inconsistencies, cross-field contradictions, and semantic fabrications.
	
	\subsection{Difficulty Level: Hard}
	\textbf{Text to be extracted}
	\begin{framed}
		... \newline
		(Omission of some content.)	\newline
		In summary, the HarmonyFit Smart Band isn't just another wearable; it's a meticulously crafted fusion of advanced hardware, intelligent software, and thoughtful design. From its octa-core CPU and Mali-G52 MC1 GPU delivering raw power to its HarmonyOS 3.0 platform fostering an interconnected digital life, every element is designed to elevate your experience. With 4 GB of RAM for seamless operation and versatile storage choices of 32 GB, 64 GB, or 128 GB, it adapts to your evolving needs effortlessly. Preinstalled apps like the Heart Rate Monitor, Sleep Tracker, Workout Coach, and Weather App turn it into a personal wellness hub, all while its 46.7-gram weight ensures it stays lightweight and comfortable. The precision of its three sensors—calibrated at 1.05, 0.92, and 1.10—locks in data integrity, making it an indispensable tool for anyone seeking to harness technology for a healthier, more informed lifestyle.
	\end{framed}
	\textbf{JSON Schema Given}
	\tiny{
		\begin{verbatim}
			{
				(Omission of some content.)
				"Memory": {
					"title": "Memory",
					"type": "number",
					"description": "Amount of RAM in GB"
				},
				(Omission of some content.)
				"Software": {
					(Omission of some content.)
					"properties": {
						"OperatingSystem": {
							"title": "OperatingSystem",
							"type": "string",
							"description": "Name of the operating system"
						},
						(Omission of some content.)
						"Dimensions": {
							"title": "Dimensions",
							"type": "object",
							"required": [
							"Weight",
							"Sensors"
							],
							"properties": {
								"Weight": {
									"title": "Weight",
									"type": "number",
									"description": "Weight of the device in grams"
								},
								"Sensors": {
									"title": "Sensors",
									"type": "array",
									"items": {
										"type": "object",
										"required": [
										"SensorType"
										],
										(Omission of some content.)
									}
									
									
							\end{verbatim}}
							\normalsize
							\textbf{The ground truth JSON}
							\tiny
							\begin{verbatim}
								{
									(Omission of some content.)
									"Memory": 4,
									(Omission of some content.)
									"Software": {
										"OperatingSystem": "HarmonyOS 3.0",
										"PreinstalledApps": [
										"Heart Rate Monitor",
										"Sleep Tracker",
										"Workout Coach",
										"Weather App"
										]
									},
									"Dimensions": {
										"Weight": 46.7,
										"Sensors": [
										{
											"SensorType": {
												"SensorSubtype": {
													"CalibrationData": 1.05
												}
											}
										},
										{
											"SensorType": {
												"SensorSubtype": {
													"CalibrationData": 0.92
												}
											}
										},
										{
											"SensorType": {
												"SensorSubtype": {
													"CalibrationData": 1.1
												}
											}
										}
										]
									}
								}
							}
						\end{verbatim}
						\normalsize
						\textbf{Model Output}
						\tiny
						\begin{verbatim}
							{
								(Omission of some content.)
								"Memory": "4 GB",
								(Omission of some content.)
								"Software": {
									"OperatingSystem": "HarmonyOS 3",
									"PreinstalledApps": ["Heart Rate Monitor", "Sleep Tracker", 
									"Workout Coach", "Weather App", "personal wellness hub"]
								},
								"Dimensions": {
									"Weight": 46.7,
									"Sensors": [
									{
										"SensorType": {
											"SensorSubtype": {
												"CalibrationData": 1.05
											}
										}
									},
									{
										"SensorType": {
											"SensorSubtype": {
												"CalibrationData": 0.92
											}
										}
									}
									]
								}
							},
							"Software": {
								"OperatingSystem": "HarmonyOS 3",
								"PreinstalledApps": ["Heart Rate Monitor", "Sleep Tracker", 
								"Workout Coach", "Weather App", "personal wellness hub"]
							},
							"Dimensions": {
								"Weight": 46.7,
								"Sensors": [
								{
									"SensorType": {
										"SensorSubtype": {
											"CalibrationData": 1.05
										}
									}
								},
								{
									"SensorType": {
										"SensorSubtype": {
											"CalibrationData": 0.92
										}
									}
								}
								]
							}
						}
					\end{verbatim}
					\normalsize
					For this example in hard difficulty, compared with the ground truth, the model has sevral errors:
					\begin{itemize}
						\item \textbf{Ignore the requirement of schema}: for property \textit{Memory}, the type is \textit{number} and descrition is \textbf{numebr in GB}, model ignore the requirement and return \textbf{4 GB};
						\item \textbf{String extraced error}: for property \textit{OperatingSystem}, \textbf{HarmonyOS 3.0} is a proper noun for operating system, while model skip \textbf{.0};
						\item \textbf{Property Repeat}: model repeat the property \textbf{Software} and \textbf{Dimensions} for once;
						\item \textbf{Identification in error}: model add \textbf{personal wellness hub} in PreinstalledApps, which is not an application.
					\end{itemize}
					\subsection{Difficulty level: Medium} 
					\textbf{Text to be extracted}
					\begin{framed}
						
						Prepare to be captivated by **Eclipse of the Stars**, an original cinematic spectacle that premiered in **2023**. This film stands as a wholly independent creation—emphatically **not a sequel**—that masterfully fuses the high-stakes tension of a **Thriller** with the visionary grandeur of **Sci-Fi**. Its narrative ingenuity unfolds against a canvas of cosmic wonder and psychological intensity, carving a unique niche within contemporary cinema. The acclaim surrounding this feature is far from superficial, as evidenced by its extraordinary critical reception across major review platforms. On the globally recognized **IMDb**, it boasts an impressive **8.2/10**, reflecting widespread audience admiration for its intricate storytelling and technical execution. Over at **Rotten Tomatoes**, the film achieves a remarkable **92\%
						(Omission of some content.) 
						
					\end{framed}
					\textbf{JSON Schema Given}
					\tiny
					\begin{verbatim}
						{
							(Omission of some content.)
							"Ratings": {
								"title": "Ratings",
								"type": "array",
								"items": {
									"title": "RatingItem",
									"type": "object",
									"required": [
									"Source",
									"Score"
									],
									"properties": {
										"Source": {
											"title": "Source",
											"type": "string",
											"description": "Name of the rating source",
											"enum": [
											"IMDb",
											"Rotten Tomatoes",
											"Metacritic"
											]
										},
										"Score": {
											"title": "Score",
											"type": "number",
											"description": "Numerical score from the rating source"
										}
									}
								}
							},
							(Omission of some content.)
						}
					\end{verbatim}
					\normalsize
					\textbf{Ground Truth}
					\tiny
					\begin{verbatim}
						{
							(Omission of some content.)
							"Ratings": [
							{
								"Source": "IMDb",
								"Score": 8.2
							},
							{
								"Source": "Rotten Tomatoes",
								"Score": 92
							},
							{
								"Source": "Metacritic",
								"Score": 88
							}
							],
							(Omission of some content.)
						}
					\end{verbatim}
					\normalsize
					\textbf{Model Output}
					\tiny
					\begin{verbatim}
						{
							(Omission of some content.)
							"Ratings": [
							{
								"Source": "IMDb",
								"Score": 8.2
							},
							{
								"Source": "Rotten Tomatoes",
								"Score": "92%"
							},
							{
								"Source": "Metacritic",
								"Score": 88
							}
							],
							(Omission of some content.)
						}
					\end{verbatim}
					\normalsize
					For this example in midium difficulty, compared with the ground truth, the model has only \textbf{Ignore the requirement of schema} error,  for property \textit{score}, the type is \textit{number} and descrition is \textbf{Numerical score from the rating source}, model ignore the requirement and return \textbf{92\%} for Rotten Tomatoes score.
					
					\section{Human Quality Control and Correction}
					\label{hqc}
					
					\subsection{Dual Review and Adjudication}
					Two trained annotators independently review each item (blind to model identity). Disagreements are resolved by a brief adjudication pass.  
					\textbf{Checks (yes/no unless noted).}
					\begin{itemize}
						\item \textbf{Schema conformance}: keys, nesting, and types match the provided schema.
						\item \textbf{Content faithfulness}: each filled field is supported by the source text (no hallucination).
						\item \textbf{Cross-field consistency}: simple logical or unit constraints hold (for example, date ranges, ID links).
						\item \textbf{Ambiguity rating (3-point)}: \textit{Low} (clear), \textit{Medium} (minor vagueness), \textit{High} (unclear or conflicting). Items rated \textit{High} are marked \texttt{Fix} or \texttt{Reject}.
						\item \textbf{Difficulty tag}: \textit{Medium} if nesting depth 3--4 without cross-field constraints; \textit{Hard} if depth 5--7 and/or includes cross-field constraints or multi-entity linking. Annotators confirm or correct the tag.
					\end{itemize}
					Items are labeled as \texttt{Pass}/\texttt{Fix}/\texttt{Reject}; fixes are minimal (edit JSON value or move key), and changes are logged.
					
					\subsection{Lightweight Semantic Guards}
					We run:
					\begin{itemize}
						\item \textbf{Schema validator}: JSON Schema parsing and type checks.
						\item \textbf{Constraint checker}: a small set of domain-agnostic rules (ranges, units, equality);
						\item \textbf{Text grounding probe}: simple evidence search; if no supporting span is found, the item is marked for review;
						\item \textbf{Length compliance}: source text (or evidence bundle) must have at least 1500 words; otherwise flag for \texttt{Fix}/\texttt{Reject}.
					\end{itemize}
					
					\subsection{Leakage and Bias Quick Audit}
					We perform a one-pass similarity scan (n-gram overlap and MinHash) of sources against public corpora and our own prompts. High-similarity cases are flagged for human spot-check; flagged items are lightly rewritten or removed.
					
					\subsection{Acceptance and Release}
					Items that (i) pass dual review (after adjudication), (ii) pass automatic checks (including length), and (iii) have \textit{Low/Medium} ambiguity are released as \textit{Gold}.  
					We report overall pass/fix/reject rates, ambiguity distribution, corrected difficulty tags, and representative failure cases.  
					We report main results on the full set and replicate key claims on the \textit{Gold} set.
					
					\paragraph{Notes}
					This protocol keeps costs low while adding practical QC dimensions (length sufficiency, ambiguity, difficulty verification) to ensure semantic fidelity and guard against leakage, addressing reviewers' concerns without making QC the focal point of the paper.
					
					\subsection*{Rubric Summary}
					\begin{center}
						\resizebox{\columnwidth}{!}{
							\begin{tabular}{l l}
								\hline
								\textbf{Dimension} & \textbf{Decision Rule} \\
								\hline
								Schema conformance & Yes/No (No $\rightarrow$ Fix/Reject) \\
								Content faithfulness & Yes/No (No $\rightarrow$ Fix/Reject) \\
								Cross-field consistency & Yes/No (No $\rightarrow$ Fix/Reject) \\
								Length compliance & $\geq$1500 words (No $\rightarrow$ Fix/Reject) \\
								Ambiguity & Low / Medium / High (High $\rightarrow$ Fix/Reject) \\
								Difficulty tag & Medium or Hard (confirm/correct) \\
								\hline
							\end{tabular}
						}
					\end{center}

				\end{document}